\documentclass[11pt, dvipsnames, DIV=10, sfdefaults=false,headings=standardclasses]{scrartcl}
\usepackage[english]{babel}

\usepackage[utf8]{inputenc}
\usepackage{CJKutf8}
\usepackage[T1]{fontenc}
\usepackage{natbib}
\usepackage{subcaption}
\usepackage{booktabs}
\usepackage{multirow}
\usepackage{url}
\usepackage{soul}
\usepackage{tikz}
\usetikzlibrary{calc}
\usetikzlibrary{decorations.pathmorphing}
\usetikzlibrary{positioning}
\usetikzlibrary{shapes}
\usetikzlibrary{arrows.meta,backgrounds,automata}
\usetikzlibrary{positioning,shapes,matrix,calc}
\tikzstyle{vertex}=[circle, draw, fill=gray!80!white,thick,scale=1.2]
\tikzstyle{edge}=[draw=black, thick,-]
\usetikzlibrary{hobby,arrows,decorations.pathmorphing,backgrounds,positioning,fit,petri,calc}
\pgfdeclarelayer{background}
\pgfsetlayers{background,main}
\addtokomafont{disposition}{\rmfamily}
\addtokomafont{descriptionlabel}{\rmfamily}

\usepackage{tikz}
\usetikzlibrary{calc}
\usetikzlibrary{decorations.pathmorphing}
\usepackage{tikz}
\usetikzlibrary{calc}
\usetikzlibrary{decorations.pathmorphing}
\usetikzlibrary{patterns,arrows,automata,positioning,decorations,shapes,calc}

\tikzstyle{normalvertex}=[circle,fill=white,draw=black]
\tikzstyle{emptyvertex}=[draw,circle,minimum size=7pt,inner sep=0pt]
\tikzstyle{tinyvertex}=[draw,circle,minimum size=3pt,inner sep=0pt]
\tikzstyle{thickedge}=[draw,gray!60,line width=1.6pt,-]

\usepackage{pgfplots}
\usepgfplotslibrary{colorbrewer}

\pgfdeclarelayer{background}
\pgfsetlayers{background,main}

\usetikzlibrary{shapes}
\usetikzlibrary{arrows.meta,backgrounds,automata, chains}
\usetikzlibrary{positioning,shapes,matrix,calc}
\tikzstyle{vertex}=[circle, draw, fill=gray!80!white,thick,scale=1.2]
\tikzstyle{edge}=[draw=black, thick,-]

\definecolor{purple}{RGB}{147,7,204}
\definecolor{blue}{RGB}{10,153,201}
\definecolor{orange}{RGB}{254,128,41}
\definecolor{pink}{RGB}{254,15,127}
\definecolor{green}{RGB}{140,211,89}

\definecolor{color1}{RGB}{254,15,127}
\definecolor{color2}{RGB}{10,153,201}
\definecolor{color3}{RGB}{194,145,62}
\definecolor{color4}{RGB}{254,128,41}
\definecolor{color5}{RGB}{254,191,185}
\definecolor{color6}{RGB}{140,211,89}
\definecolor{color7}{RGB}{245,221,66}

\tikzstyle{vertex} = [fill,shape=circle,node distance=80pt]
\tikzstyle{edge} = [fill,opacity=.5,fill opacity=.0.05,line cap=round, line join=round, line width=5pt]
\tikzstyle{elabel} =  [fill,shape=circle,node distance=30pt]
\pgfdeclarelayer{background}
\pgfsetlayers{background,main}

\usetikzlibrary{calc}
\usetikzlibrary{decorations.pathmorphing}
\usetikzlibrary{decorations.text}
\usetikzlibrary{positioning, arrows.meta, fit}
\usetikzlibrary{shapes}
\usetikzlibrary{arrows.meta,backgrounds,automata}
\usetikzlibrary{matrix,calc,bending}
\tikzstyle{vertex}=[circle, draw, fill=gray!80!white,thick,scale=1.2]
\tikzstyle{edge}=[draw=black, thick,-]
\usetikzlibrary{hobby,arrows,decorations.pathmorphing,backgrounds,positioning,fit,petri,calc}
\tikzstyle{vertex}=[anchor=center, circle, fill=gray, inner sep=1.25]

\tikzset{
	invisible/.style={opacity=0,text opacity=0},
	visible on/.style={alt=#1{}{invisible}},
	alt/.code args={<#1>#2#3}{%
		\alt<#1>{\pgfkeysalso{#2}}{\pgfkeysalso{#3}} %
	},
}

\usetikzlibrary{calc, math, arrows.meta}

\pgfdeclarelayer{bg}    
\pgfsetlayers{bg,main}

\usetikzlibrary{arrows,positioning} 
\tikzset{
	>=stealth',
	punkt/.style={
		rectangle,
		rounded corners,
		draw=black, very thick,
		minimum height=2em,
		text centered},
	pil/.style={
		->,
		thick,
		shorten <=2pt,
		shorten >=2pt,}
}

\usepackage{paralist}
\usepackage[stable]{footmisc}
\usepackage{adjustbox}

\usepackage{ifthen}
\newcommand{\CC}[1][]{$\text{C\hspace{-.25ex}}^{_{_{_{++}}}}
\ifthenelse{\equal{#1}{}}{}{\text{\hspace{-.625ex}#1}}$}

\usepackage{bm}
\usepackage{bbm}
\usepackage{amsmath}
\usepackage{amssymb}
\usepackage{amsthm}
\usepackage{amsfonts}
\usepackage{thmtools}		
\usepackage{mleftright}
\usepackage{stmaryrd}
\usepackage{nicefrac}
\usepackage{algorithm}
\usepackage{algorithmicx}
\usepackage[noend]{algpseudocode}

\let\originalleft\left
\let\originalright\right
\renewcommand{\left}{\mathopen{}\mathclose\bgroup\originalleft}
\renewcommand{\right}{\aftergroup\egroup\originalright}

\usepackage{pifont}

\usepackage{enumitem}
\setlist[enumerate]{itemsep=0.2ex, topsep=0.5\topsep}
\setlist[description]{itemsep=0.2ex, topsep=0.5\topsep}
\setlist[itemize]{itemsep=0.2ex, topsep=0.5\topsep}

\makeatletter
\def\thmt@refnamewithcomma #1#2#3,#4,#5\@nil{%
\@xa\def\csname\thmt@envname #1utorefname\endcsname{#3}%
\ifcsname #2refname\endcsname
\csname #2refname\expandafter\endcsname\expandafter{\thmt@envname}{#3}{#4}%
\fi
}
\makeatother

\usepackage[pagebackref,
pdfa,
hidelinks,
pdftex, 
pdfdisplaydoctitle,
pdfpagelabels,
pdfauthor={Christopher Morris},
pdftitle={Future Directions in Foundations of Graph
Machine Learning},
pdfsubject={},
pdfkeywords={GNNs, generalization, optimization, expressivity},
pdfproducer={Latex with the hyperref package},
pdfcreator={pdflatex}
]{hyperref}

\usepackage[capitalise,noabbrev]{cleveref}   
\theoremstyle{definition}

\usepackage{thm-restate}
\usepackage[mathic=true]{mathtools}
\usepackage{fixmath}
\usepackage{siunitx}

\usepackage[activate={true,nocompatibility},final,kerning=true]{microtype}
\usepackage{ellipsis}

\usepackage[scaled=0.86]{helvet}
\usepackage{newtxtext}
\usepackage{newtxmath}

\usepackage[auth-lg]{authblk}

\newcommand{\wlone}{$1$\textrm{-}\textsf{WL}}
\newcommand{\kwl}{$k$\textrm{-}\textsf{WL}}

\usepackage{xspace}

\newcommand{\new}[1]{\emph{#1}}

\title{\huge \normalfont\textbf{Future Directions in the Theory of Graph Machine Learning}}

\author[1]{Christopher Morris}
\author[2]{Fabrizio Frasca}
\author[2]{Nadav Dym}
\author[2,3]{Haggai Maron}
\author[4]{İsmail İlkan Ceylan}
\author[2]{Ron Levie}
\author[5]{Derek Lim}
\author[4]{Michael Bronstein}
\author[1]{Martin Grohe}
\author[5,6]{Stefanie Jegelka}

\affil[1]{RWTH Aachen University}
\affil[2]{Technion - Israel Institute of Technology}
\affil[3]{NVIDIA Research}
\affil[4]{University of Oxford}
\affil[5]{MIT}
\affil[6]{TU Munich}
\date{\vspace{-30pt}}

\recalctypearea

\pgfplotsset{compat=1.18}
\begin{document}

\maketitle
\begin{abstract}
	Machine learning on graphs, especially using graph neural networks (GNNs), has seen a surge in interest due to the wide availability of graph data across a broad spectrum of disciplines, from life to social and engineering sciences. Despite their practical success, our theoretical understanding of the properties of GNNs remains highly incomplete. Recent theoretical advancements primarily focus on elucidating the coarse-grained expressive power of GNNs, predominantly employing combinatorial techniques. However, these studies do not perfectly align with practice, particularly in understanding the generalization behavior of GNNs when trained with stochastic first-order optimization techniques. In this position paper, we argue that the graph machine learning community needs to shift its attention to developing a balanced theory of graph machine learning, focusing on a more thorough understanding of the interplay of expressive power, generalization, and optimization.
\end{abstract}

\section{Introduction}
Graphs serve as powerful mathematical representations, adept at capturing intricate interactions among entities across a spectrum of disciplines, spanning from life~\citep{Won+2023} to social sciences~\citep{Eas+2010} and optimization~\citep{Cap+2021}. This diversity underscores the critical demand for specialized machine-learning methods that extract valuable patterns from complex graph data.

Hence, in recent years, neural networks capable of handling graph-structured data received a lot of attention in the machine learning community, especially \new{messsage-passing graph neural networks} (MPNNs)~\citep{Gil+2017,Sca+2009}.\footnote{We use the term MPNNs to refer to graph-machine learning architectures that fit into the framework of~\citet{Gil+2017} and use the term GNNs in a broader sense, i.e., all neural network architectures capable of handling graph-structured inputs.} Nowadays, MPNNs, or, more generally, GNNs, are among the most prominent topics at top-tier machine learning conferences,\footnote{\url{http://tinyurl.com/mpn89vju}} and have showcased promising outcomes across diverse domains, including breakthroughs in discovering new antibiotics~\citep{Sto+2020,Won+2023}.

\begin{figure}[t]
	\begin{center}
		\scalebox{0.65}{
			\begin{tikzpicture}
				\node[] (dummy) {};

				\node[punkt,left= of dummy, fill=lightgray, minimum width=80pt,draw=black] (a) {\textsf{Expressivity}};
				\node[punkt,below= of a, fill=lightgray,minimum width=80pt,draw=black] (b) {\textsf{Graph Structure $\sim$ Data}};
				\node[punkt,below left= of b, fill=lightgray,minimum width=80pt,draw=black] (c) {\textsf{Generalization}};
				\node[punkt,below right = of b, fill=lightgray,minimum width=80pt,draw=black] (d) {\textsf{Optimization}};

				\node[punkt, fill=ForestGreen!50,minimum width=105pt,draw=black] at (-11.0,0.0) (aa) {\textsf{Architectural choices}};

				\node[punkt, fill=ForestGreen!50,minimum width=105pt,draw=black] at (-11.0,-1.75) (bb) {\textsf{Model parameters}};

				\node[punkt, fill=ForestGreen!50,minimum width=105pt,draw=black] at (-11.0,-3.60) (bb) {\textsf{Graph classes}};

				\node[punkt, fill=BurntOrange!50,minimum width=80pt,draw=black] at (7.5,-1.75) (e) {\textsf{Applications}};

				\draw [>={Stealth[length=8pt,round]}, <-, thick] (a) to  (b);
				\draw [>={Stealth[length=8pt,round]},<-, thick, bend left] (c) to  (b.west);
				\draw [>={Stealth[length=8pt,round]},<-, thick, bend right] (d) to  (b.east);
				\def\myshift#1{\raisebox{0.6ex}}
				\draw [>={Stealth[length=8pt,round]},<->, thick, bend left, postaction={decorate,decoration={text along path,text align=center,text={|\sffamily\myshift|Influences}}}] (c.north) to   (a.west);
				\draw [>={Stealth[length=8pt,round]},<->, thick, bend left, postaction={decorate,decoration={text along path,text align=center,text={|\sffamily\myshift|Influences}}}] (a.east) to  (d.north);

				\draw [>={Stealth[length=8pt,round]},<->, thick, postaction={decorate,decoration={text along path,text align=center,text={|\sffamily\myshift|Influences}}}]   (c) to  (d);

				\begin{pgfonlayer}{bg}
					\path[draw=lightgray!20, line width=4mm,<->, >=stealth, shorten >=7pt, shorten <=7pt]    (a.east) to [bend left]    (e.north);
					\path[draw=lightgray!20, line width=4mm, <->, >=stealth,  shorten >=5pt, shorten <=7pt]    (d.north) to [bend left]    (e.west);
					\path[draw=lightgray!20, line width=4mm,<->, >=stealth,  shorten >=10pt, shorten <=20pt]    (c.south) to [bend right]    (e.south);

				\end{pgfonlayer}

			\end{tikzpicture}}
		\vspace{-25pt}
	\end{center}
	\caption{\label{fig:overview} Interactions of the four challenges within graph machine learning: Fine-grained \emph{expressivity}, \emph{generalization}, \emph{optimization}, \emph{applications}, and their interactions. The green boxes \emph{architectural choices} (hyperparameter and other design choices like normalization layers), \emph{model parameters}, and \emph{graph classes} (different types of graphs) represent aspects of all four challenges.}
\end{figure}
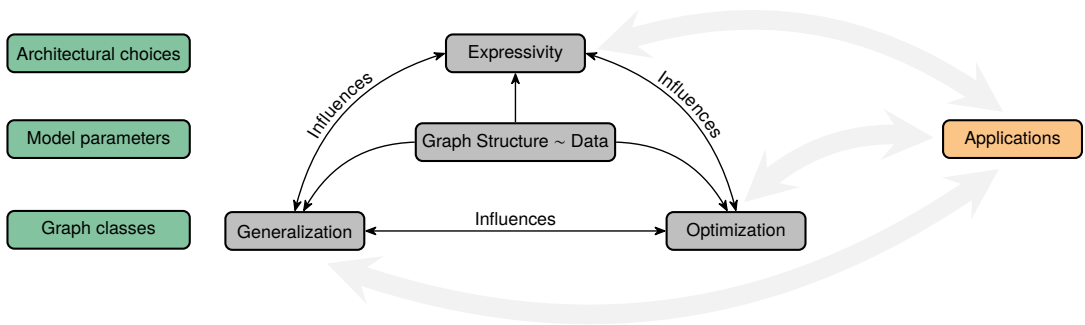

While GNNs are successful in practice and are making real-world impact, their theoretical properties are understood to a lesser extent. That is, only GNNs' expressive power, i.e., their ability to separate graphs and express functions over graphs is understood to some extent~\citep{Azi+2020,geerts2022,Mor+2019,Mor+2022,Xu+2018b}. However, most current analyses heavily rely on combinatorial techniques, such as the \new{$1$-dimensional Weisfeiler--Leman algorithm} (\wlone), a well-studied heuristic for the graph isomorphism problem~\citep{Gro2017,Wei+1968,Wei+1976}. While the graph isomorphism perspective has helped the community understand GNNs' ultimate limitations in capturing graph structure, it is inherently binary. For example, it does not give insights into the degree of similarity between two given graphs, prohibiting a more fine-grained analysis. While some recent works~\citep{boeker2023finegrained,chen2022weisfeiler} aim at a more fine-grained analysis, they still have strong limitations, such as not considering continuous node and edge features. A second limitation of current GNN expressivity results is that they are fairly specific. They are tailored to particular classes of GNNs, overlooking practically relevant architectural choices. 

While understanding {\color{black}MPNNs'} and related architectures, e.g., graph transformers~\citep{Mue+2023}, expressive power is vital, understanding when such architectures generalize to unseen graphs and how to find parameter assignments that allow so is equally important. However, in MPNNs and GNNs, the essential aspects of \new{generalization} and \new{optimization} are severely understudied. The few existing works that study {\color{black}MPNNs'} generalization properties, e.g., \citet{Gar+2020,Lia+2021,Mas+2022,Sca+2018}, study {\color{black}MPNNs'} generalization properties via VC (Vapnik-Chervonenkis) dimension theory or related formalisms and derive generalization bounds that depend on high-level graph parameters such as the maximum degree or number of nodes. In addition, all current works are based on variants of classical uniform generalization bounds. This entails large constants in the bound and a description of a classical bias-variance curve that does not describe the typical reality of deep learning on graphs, in which higher complexity and more expressive models often generalize better.

While some initial works make progress towards studying {\color{black}GNNs'} optimization aspects, i.e., the dynamics of \emph{stochastic gradient descent} (SGD) to adapt their parameters, they still make strong assumptions, such as the use of linear activation functions~\citep{Xu+2021}, unrealistic learning scenarios~\citep{Du+2019b}, or neglecting the influence of the graph structure~\citep{Tan+2023}; see also \citep{bechler2023graph}.

Overall, we argue that the following challenges persist in graph machine learning.

\begin{description}
	\item[Expressivity] The vast majority of existing results on {\color{black}GNNs'} expressive power are coarse-grained and focus on specific architectures.
		Additionally, guidelines for choosing between highly-expressive GNNs are needed.

	\item[Generalization] Besides expressivity, it is vital to choose models based on a training dataset so that it generalizes to unseen data. Current works studying GNNs' generalization properties often rely on many simplifying assumptions, e.g., not considering graph structure or optimization.

	\item[Optimization] We must ensure that GNNs trained with SGD converge to assignments leading to expressive models that generalize well and understand the role of the architecture and graph structure.

	\item[Applications] Current theoretical results are often loosely aligned with practical assumptions and needs of application domains. Hence, developing graph machine learning theory aligned with practical requirements is crucial.
\end{description}

We argue that graph machine learning needs a nuanced theory to develop an in-depth understanding of the various architectural choices that govern {\color{black}GNNs'} expressive power (challenge 1), generalization properties (challenge 2), and optimization dynamics (challenge 3), as well as the \emph{interplay} of these aspects. Moreover, theoretical research within the graph machine learning communities must be aligned with domain experts' practical needs (challenge 4). Thereto, we propose concrete challenges to address these requirements; see~\cref{fig:overview} for a high-level overview of the interactions of the four challenges in this position paper.

\section{Expressive Power of GNNs}\label{sec:expressivity}
Most of the expressiveness studies examine the ability of GNNs to assign distinct values to non-isomorphic graphs, or in other words, the ability of GNNs to separate different graphs. Moreover, the ability to separate graphs is directly connected to {\color{black} MPNNs'} ability to approximate (continuous, permutation-invariant) functions over graphs~\citep{chen2019}. The seminal works of~\citet{Mor+2019,Xu+2018b} showed that MPNNs' ability to separate graphs is equivalent to the \emph{limited} separation power of the \wlone{} algorithm for the graph isomorphism problem~\citep{Wei+1968}. These works inspired the study of \new{expressive GNNs}, whose expressive power surpasses that of the \wlone{} test, typically at the cost of more computational resources. For example, \citet{Mor+2019,Mar+2019c} derived architectures equivalent to the more powerful $k$-dimensional Weisfeiler--Leman algorithm (\kwl). Many other types of expressive GNNs were proposed in the literature, e.g., by utilizing random features~\citep{Abb+2020}, subgraph counts \citep{Bou+2020}, or by employing sets of subgraphs~\citep{Bev+2021,Cot+2021,Qia+2022,Fra+2022}. See the following surveys for a more thorough review: \cite{jegelka2022theory,li2022expressive,Mor+2022,sato2020survey,zhang2023expressive}. Several recent works developed tools to analyze the expressive power of GNN architectures or construct hierarchies thereof \citep{geerts2022,zhang2023complete}. Alternative expressive power measures have been proposed in the past several years, for example, measuring the degree of invariant or equivariant polynomials an architecture can represent \citep{puny2023equivariant}, or measures based on the ability to compute the count of substructures \citep{chen2020can} or other graph properties \citep{zhang2023rethinking} and to mix information from different nodes \citep{di2023does}.

\subsection{Challenges}

Based on the above, we identify the following challenges.

\paragraph{Challenge II.1: From combinatorial to geometric expressiveness results.} The common results on the expressive power of GNNs are \new{coarse-grained.} Since these tools are rooted in graph isomorphism testing formalism, and the graph isomorphism problem inherently involves binary outcomes, they fall short of offering insights into the {\em degree of similarity} between graphs. This omission potentially limits the scope of a nuanced analysis of GNNs. For example, \citet{Mar+2019c,Mor+2019,Xu+2018b} have independently devised \wlone-expressive GNNs, distinguishing the same set of graphs. However, their empirical performance in practical situations is not identical, and which one is better varies across tasks.

To guide practitioners in selecting these maximally expressive MPNNs, we need to consider the {\em geometry of the space of graphs} induced by the geometry of the feature space produced by the MPNNs. Note that an MPNN is a function that maps graphs to features in some Euclidean space, and the Euclidean metric in the feature space can be pulled-back to a (pseudo) metric of graphs. The choice of architecture determines this graph metric and which graphs can be separated by the architecture and how easily. Understanding this \new{fine-grained expressivity} could lead to a more systematic approach to designing GNN architectures. In addition, using such a ``continuous'' notion of graph similarity will also lead to results akin to {\em geometric stability} in geometric deep learning \citep{bruna2013invariant}, potentially related to generalization; see~\cref{sec:generalization}.

While~\citet{boeker2023finegrained} took a first step towards a non-binary class expressivity and identified several well-known metrics in the graph space, which are topologically equivalent to the Euclidean metric in the feature space; the results only consider graphs without continuous node or edge features, use specific aggregation functions, and do not offer an explicit way to bound graph distances by feature distances and vice-versa. As an initial step, \cite{chuang22tmd} derived a pseudometric on graphs with continuous attributes based on the computation trees of MPNNs. Additional related works are on transferability and convergence of GNNs, which only give one side of a topological equivalency, namely, convergence of graphs lead to convergence of features \citep{Levie+2022,keriven+2022,Ruisz+2020}.

We propose developing \new{fine-grained expressivity} results, namely metric equivalencies between explicit graph metrics and feature metrics for GNNs on graphs with features. An ideal result would derive a bi-Lipschitz correspondence between such metrics. Note that \citet{Lev+2023} gave one side of the Lipschitz inequality for graphs with features, namely, the Euclidean feature distance is bounded by the graph cut-distance for any MPNN. Lastly, fine-grained expressivity can lead to \emph{optimal universal approximation theorems}. Universal approximation results based on the Stone--Weierstrass theorem require a better understanding of the topology of the space of graphs. The finer the topology we consider on the input space, the more points are ``far apart'' in this space, and therefore, the harder it is to separate far-apart points using functions. Hence, optimal universal approximation theorems should find the finest topology in which MPNNs separate points.

\paragraph{Challenge II.2: Towards understanding expressiveness for all practical architectures.} A second limitation of current GNN expressivity results is that they are very \new{specific}. They are tailored to specific classes of GNNs, overlooking practical and relevant architectural choices. More specifically, the equivalence of MPNNs and the \wlone~ is obtained for particular choices of MPNN architectures~\citep{Mor+2019,Xu+2018b}. Therefore, there is a need to understand the effect of various architectural decisions made in practice, such as different activations, aggregation, normalization, and pooling, on the expressive power of MPNN architectures. Examples of results in this vein are the expressive advantage of sum pooling over mean and max pooling \citep{Xu+2018b} and the expressive advantage of analytic activations over piecewise linear activations \citep{amir2023neural}.

Apart from MPNNs, a comprehensive understanding of expressivity is especially lacking for graph transformers (GTs). The effectiveness of GTs heavily relies on incorporating structural and positional encodings~\citep{Mue+2023}. These encodings introduce information about the underlying graph structure into the transformer architecture, which is inherently designed without an awareness of graph structures. However, it is still largely unclear how these encodings influence an architecture's ability to capture graph structure; see some preliminary results in \citet{Lim+2023,Lim+2023a,zhu2023structural, zhang2023rethinking}, and also \citet{cai2023connection,rosenbluth2024distinguished} for some connections between MPNNs and GTs. However, it remains unclear how to abstract the various architectural choices and engineering tricks via a mathematical model, allowing for a detailed mathematical analysis of GT's potential benefits over MPNNs.

\paragraph{Challenge II.3: Towards uniform expressiveness results.}
Most GNN expressive power results in the literature either do not quantify or only give loose bounds on the size of the GNN required (in terms of number of parameters, width, or depth) to compute different functions on graphs, e.g.,~\citet{aamand2022exponentially,amir2023neural,Mor+2019}. Moreover, most expressivity results are \emph{non-uniform}, i.e., they depend on the graph size. Further study on this could have several benefits. First, the size of a GNN required for different tasks is related to its generalization ability. Secondly, the size of GNNs required for expressing certain functions could help compare different GNN architectures that have similar or the same expressive power; for instance, for two MPNN architectures that are \wlone{} equivalent, one may prefer the one that requires fewer parameters for distinguishing \wlone-distinguishable graphs. Moreover, deriving uniform expressivity results could shed light on GNNs' ability to generalize to larger graphs, as seen during training.
\citet{barcelo2020} achieves a result that is uniform, i.e., one model parametrization can express the target function on \emph{all} graphs; see also \citet{Grohe21logic} for discussions. However, the result hinges on the use of logical classifiers. In addition, \citet{Grohe23descriptive} proves that all functions computable by MPNNs are expressible in the logic $FO^2+C$, leading to an exact characterization of the functions that are computable by MPNNs of polynomial size and bounded depth in terms of logic and also standard computational complexity classes.

\paragraph{Challenge II.4: Towards expressiveness on relevant classes of graphs.} Most expressive GNNs have been designed to attain high separation power on the general family of all possible graphs. However, practical applications typically target specific classes of graphs for which tailored expressiveness results may be obtained or are otherwise known. For example, most molecules are in the family of \emph{planar} graphs~\citep{SIMMONS1981287}, on which the graph isomorphism problem can be solved more efficiently by specialized algorithms~\citep{hopcroft1974linear}. Because of this, we advocate addressing the challenge of developing the study of expressive GNN architectures targeting relevant graph classes of interest, following pioneering works by~\citet{dimitrov2023plane} on planar graphs, and by~\citet{bause2023maximally} on outer-planar graphs. Besides atomistic systems, other examples include bipartite or tripartite graphs stemming from optimization problems~\citep{Cap+2021,Qia+2024} or recommender systems~\citep{Wu+2023}. 

We suggest identifying a taxonomy of families of graphs that hold particular importance in practical applications to re-evaluate the expressive power of known architectures about these and to derive optimal lower-bound intricacy figures for maximally expressive approaches on these classes; see Challenge II.5. These analyses would support the design of architectures, which, on specific graph families, can attain better complexity-expressivity tradeoffs than more general GNNs.

\paragraph{Challenge II.5: Towards a formal trade-off between expressive power and computational cost.}
Another challenge concerning expressivity is exploring the realm of \emph{expressive GNNs}. While there are many works proposing GNNs whose expressivity surpasses the \wlone{} test, the GNN community needs principled methods of navigating this vast collection of architectures by providing guidelines for how expressive a GNN architecture needs to be to address a given task \citep{di2023does}. In addition, in practice, we observe a trade-off between computational complexity and expressive power, where more expressive architectures typically come with a higher computational complexity. There is a need to quantify this trade-off formally. For example, there is a lack of theoretical knowledge to address questions such as: given specific expressive power requirements (e.g., \wlone, or \kwl, counting specific substructures), what are the lower bounds of the time and space complexity for a GNN model with the requisite representational capacity? \citet{tahmasebi2023power} recently obtained preliminary results in this direction. The answer to this question could guarantee that the GNN architectures we currently have are optimal in terms of complexity or, conversely, point towards possible improvements that can lead to GNNs with identical expressivity properties and improved complexity.

\paragraph{Challenge II.6: Towards linking architecture, task, and graph structure.}
In GNNs, the graph is part of the input and the computational device. At times, the input graph might not be ideally suited for message passing, leading to phenomena such as over-squashing \citep{Alon2020}. Characterizing the graph properties that lead to such phenomena, e.g., through curvature \citep{Top+2021} or diffusion distances \citep{di2023onover}, can lead to more principled approaches for decoupling the computational and input graphs in MPNNs through {\em rewiring} \citep{barbero2023locality,Qia+2023b}, or more efficient sparse GTs. A better characterization of the tasks that a specific MPNN architecture can implement on a particular graph, such as in the recent work of \cite{di2023does}, can also lead to a novel type of expressiveness results.

\section{Generalization Properties of GNNs}\label{sec:generalization}

The few existing works that study MPNNs' generalization properties, e.g.,~\citet{Gar+2020,Lia+2021,Mas+2022,Sca+2018}, study their generalization properties via VC dimension theory or related formalisms. However, the bounds derived in such works usually depend only on relatively generic graph parameters, such as the maximum degree or number of nodes, ignoring more expressive graph parameters. While~\citet{wlvc} recently established a tight connection between the expressivity of the \wlone{} and MPNNs' VC dimension, the analysis is restricted to a particular MPNN architecture using sum aggregation; see also~\cref{sec:expressivity}. Hence, it does not allow for fine-grained analysis, e.g., considering different architectural choices and more quantitative analysis in terms of numbers of parameters or depth. An additional important factor is that the above works assume that the train and test sets are sampled from the same distribution. Hence, they reveal little about MPNNs' ability to generalize out-of-distribution, e.g., generalizing to larger graphs than found in the training set. Furthermore, no theoretical understanding exists of the generalization abilities of more expressive GNN architectures and how they are influenced by graph structure.

\subsection{Challenges}

To advance the study of GNNs' generalization ability, we suggest to address the following challenges.

\paragraph{Challenge III.1: Understanding the influence of expressiveness and architectural choices on generalization.}
Although classical learning theory~\citep{wlvc} would suggest worse generalization results for higher expressiveness, more expressive GNN architectures often achieve better generalization performance. While the underlying reasons for these improvements are not fully clear, this phenomenon hints at the important role played by particular inductive biases and their interplay with the data distribution at hand~\citep{Bou+2020}. We thus propose to establish precise conditions under which increased expressive power, jointly with architectural choices, leads to enhanced generalization. \citet{Fra+2024} made recent progress in that direction, focusing on linearly separable data. 

We propose to analyze how expressivity transforms the feature space to facilitate better generalization, for example, by deriving conditions on the underlying data distribution such that more expressive power leads to better predictive performance. At the same time, it is pivotal to unravel the influence of different design choices on GNNs' generalization abilities, including, e.g., aggregation functions, skip connections, or normalization layers. For example, an essential aspect would be understanding if certain aggregation functions exhibit superior generalization and offer advantages in making predictions on larger graphs beyond the training set and for which tasks. A first step in this direction was taken by \citet{xu20reasoning,xu2021how}. Related to this, we deem it essential to investigate architectural paradigms beyond message-passing, such as graph transformers~\citep{Mue+2023}. In this sense, a relevant endeavor would be to decipher how the attention mechanism contributes to improved generalization compared to MPNNs, the influence of various structural and positional encodings on this aspect, and how they influence generalization bounds. Building on advancements in understanding the expressive power of structural encodings, see~\cref{sec:expressivity}, we propose to understand if certain encodings offer better generalization properties and how different encodings influence the sample complexity.

\paragraph{Challenge III.2: Understanding the impact of graph structure on generalization and its interplay geometry.} 
This challenge asks how graph structure influences an architecture's generalization properties and how this is linked to an architecture's expressive power and the feature space's geometry. For example, it is essential to understand if graph properties such as sparsity influence generalization and how to integrate them into generalization bounds.
Moreover, considering practical, relevant graph classes such as planar or bipartite graphs, it is important to precisely understand whether GNNs operating on certain graph classes offer better generalization than others. Further, leveraging results on the fine-grained geometry of the feature space outlined in~\cref{sec:expressivity} possibly allows to derive tighter generalization bounds. While \citet{Lev+2023} took a first step in this direction, they did not consider the coarsest topology in the space of graphs equivalent to the geometry of the feature space, and hence their results are suboptimal.

\paragraph{Challenge III.3: Develop a theory of data augmentation.}
Given the cost and scarcity of labeled training data in many practical graph machine learning scenarios, several recent works (e.g., as surveyed in \citet{ding2022data}) suggested leveraging \new{data augmentation} techniques (i.e., enhancing the training dataset with additional labeled data samples) to increase the training data size effectively. Importantly, there is little theoretical understanding, existing work, e.g., uses graphons~\citep{Han+2022}, and relatively little practical evidence regarding which data augmentation schemes are helpful for specific graph distributions and learning tasks. Recently, graph data augmentation strategies have seen application in mitigating covariate distribution shifts~\citep{sui2023unleashing} and in rationalizing GNN predictions~\citep{liu2024rationalizing,wu2022discovering}. Unfortunately, however, compared to the widespread beneficial effect data augmentation has on learning with other popular data modalities---such as images---the effectiveness of data augmentation in graph machine learning still lags behind. Hence, a critical challenge is the development of a general theory of data augmentation for graph-structured data. Hence, we suggest deriving a better understanding of how much we can perturb graph data without distorting the underlying data-generating distribution, e.g., leveraging recent advancements in principled graph similarity~\citep{Ger+2022}.

\paragraph{Challenge III.4: Understanding and improving extrapolation, especially to larger graphs.}
Extrapolation to graphs sampled from a different distribution than the training distribution is crucial in many applications. For instance, extrapolation to larger graphs is especially important in tasks where labels are much more expensive for larger graphs, such as combinatorial optimization applications, where solvers are slow on large instances~\citep{Cap+2021}. It has been shown both theoretically and empirically that GNNs can fail on graphs that are larger than, or in general have a different structure than, the graphs that they are trained on~\citep{yehudai2021local,velivckovic2022clrs,xu2021how,zhou2022ood}. Hence, we need to better understand under which conditions on data distributions, models, and optimization out-of-distribution generalization is possible. From the perspective of model and optimization, \citet{xu2021how} take a first step in showing how restrictions on the architecture can enable extrapolation to different graph structures. In addition, initial results \citep{adamday2023zeroone} within the Erdős–Rényi model, leveraging results from finite model theory, show that MPNNs eventually become independent of their inputs as the graph grows. Another perspective is to characterize theoretical conditions on the structure of the graphs in training and test sets. In transferability and convergence analysis, MPNNs are shown to be transferable between different graphs that are sampled from the same random graph model \citep{keriven+2022,le23transfer,Levie+2022,Ruisz+2020}, but we still need to better understand the scope of practical situations in which these conditions hold. More generally, we still need to analyze the entire landscape of conditions for extrapolation in practical situations, taking into account data, task, architecture, and other inductive biases, e.g.\ from optimization, for a more general and actionable picture of when GNNs and graph transformers extrapolate to different kinds of data distributions, and develop new methods for improving GNN extrapolation ability. A related question is to estimate the reliability of a model on new distributions in a way consistent with empirical behavior. This question relates to understanding the geometry of graph representations (see also Challenges II.1 and III.2) and relevant divergences of distributions that reflect GNN stability. A first work in this direction is \citet{chuang22tmd}. Answering these questions would also be a basis to build a better understanding of transfer learning for graphs.

\section{Optimization Dynamics of GNNs}\label{sec:optimization}

Only some works directly focus on the theoretical understanding of gradient descent-based learning for GNNs. These works typically make substantial simplifying assumptions. For example, \citet{Xu+2021} showed global convergence of GNNs with \emph{linear activations}. In contrast, \citet{Du+2019b} showed global convergence for the neural tangent kernel obtained as the infinite width limit of GNNs, simplifying practical scenarios. In addition, the analysis in \citet{Tan+2023} does not consider graph structure.

In contrast, experimental and theoretical evidence suggests that the GNN optimization procedure may only be optimal in some settings. In several papers \citep{boeker2023finegrained,huang2022you}, GNNs with random parameters or graph kernels with handcrafted features outperform trained GNNs. Furthermore, GNNs' performance often deteriorates as depth increases, a problem frequently attributed to phenomena such as over-smoothing, over-squashing, and graph bottlenecks. While many normalization techniques have been suggested to avoid these issues successfully, they often are not accompanied by a substantial performance gain \citep{rusch2023survey}. Additional evidence of the limitation of the GNN optimization procedure is the recent work of \citet{bechler2023graph}, which shows that GNNs perform suboptimally on learning tasks that can be solved only using node features, despite the theoretical ability of GNNs to exploit these features only and disregard the graph structure.

\subsection{Challenges}

Based on the above, we identify the following challenges.

\paragraph{Challenge IV.1: Towards guarantees for convergence quality and rate.}
One important step towards better understanding the learning of graph representation models is to improve existing analysis of convergence by considerably weakening the strong linearity assumptions used in existing work \citep{Du+2019b,Tan+2023,Xu+2021}. The convergence guarantees we are concerned with are both guarantees on the rate of convergence and guarantees on the quality of convergence, as well as a characterization of what the dynamics converge to, e.g., a global or low-loss local minimum or basin of attraction and other properties, including relations to generalization, see~\cref{sec:generalization}, e.g., via certain biases for simplicity. We may also ask whether GNN training exhibits phenomena that have been observed in other neural networks, such as the edge of stability or convergence not to a point but an invariant measure~\citep{cohen21edge,chandramoorthy22,lobacheva21periodic,zhang22invariant}.

\paragraph{Challenge IV.2: Towards understanding the influence of architectural choice on GNNs' optimization.} Here, the primary objective is to unravel the impact of various architectural choices, such as aggregation functions or normalization layers, on the properties of GNNs trained with SGD. The goal is to discern how these architectural choices shape the convergence properties of GNNs, explicitly investigating whether SGD can lead to parameter assignments that possess favorable attributes for generalization. This can be accomplished by adapting recent advancements in understanding the convergence properties of SGD on standard feed-forward neural networks~, e.g., \citet{Du+2018,Du+2019,Du+2023} and applying these insights to MPNNs. This adaptation may involve interpreting MPNN computation through the recursive unrolling of neighborhood structures and comprehending how graph structure can be incorporated into these results. This exploration should include GTs; insights from MPNNs can potentially be transferred, with modifications, to GTs. We need to investigate whether GTs exhibit distinct properties when trained with SGD. For instance, we must explore the potential significant influence of attention mechanisms within GTs.

\paragraph{Challenge IV.3: Towards understanding the influence of the graph structure on GNNs' optimization.} In other realms of deep learning, neural networks typically adhere to a consistent structure, resulting in the limited impact of their structure on the convergence properties of SGD. However, for GNNs, the neural networks' architecture is directly shaped by the underlying data distribution, i.e., dictated by the graphs' connectivity. Consequently, this challenge focuses on unraveling the intricacies of how graph structure, encompassing factors such as sparsity or specific graph classes, affects the convergence properties of SGD. In addition, we seek to understand how parameters controlling graph structures, e.g., the number of symmetries, can be incorporated into convergence guarantees. Given that GTs integrate graph structure through structural and positional encodings, it is also interesting to understand the interplay between these encodings and the convergence behavior of SGD, as well as the impact of choosing different encodings on the convergence dynamics.

\paragraph{Challenge IV.4: Towards harnessing the power of depth.}  For convolutional neural networks, successful training of deeper neural networks via normalization techniques, such as batch normalization or skip connections, have led to significant performance improvements~\citep{he2016deep}. Several works, e.g.,~\citet{PairNorm}, have developed analogous normalizations for GNNs, and in some cases, complex combinations of these techniques do lead to improved performance~\citep{li2021training}. Nevertheless, these solutions have not consistently demonstrated gains significant enough to persuade the community of their fundamental architectural role, as is arguably the case for their counterparts in other modalities.  We suggest approaching this challenge by developing toy models of graph machine-learning problems that can only be solved by deep GNNs and identifying training mechanisms that will be guaranteed to lead to successful learning. As a second step, the obtained training techniques can be applied to real-world tasks, emphasizing tasks where deeper GNNs are expected to be beneficial, such as the ``long-range graph benchmark''~\citep{longrange}.

\paragraph{Challenge IV.5: Defeating randomness.}
The existence of instances where graph features produced by GNNs with random weights perform on par with trained GNNs~\citep{böker2023finegrained,huang2022you} suggests that optimization of GNNs sometimes does not lead to significant improvement over the initial solution. This finding is coherent with that MPNNs of moderate size, with random weights, attain their maximal separation power~\citep{amir2023neural,aamand2022exponentially}. Nonetheless, it seems plausible that optimizing the GNNs' weights should lead to better graph features and more successful learning by SGD. This challenge aims to find mathematically precise explanations and models where the optimal GNN parameters are significantly better for learning than the average GNN parameters and suggest ways to successfully learn these optimal GNN parameters; see also Challenge IV.1.

\section{Connecting Theory with Practice}

While addressing the above-outlined challenges regarding the GNNs' expressive power, generalization abilities, and optimization dynamics is essential to push GNN theory forward, we also want to stress the need to align such theory with practical needs.

For example, currently, proposed expressive architectures, such as higher-order GNNs aligned with the \kwl~\citep{Mor+2022}, are rarely employed by domain experts, e.g., in molecular property prediction~\citep{Duv+2023}. In addition, theoretical results that have been derived so far often make unrealistic assumptions, ignoring practical needs such as continuous node and edge features. Moreover, results investigating the generalization properties of GNNs based on VC dimension theory and related concepts, e.g., \citet{Gar+2020,wlvc}, result in large sample complexities, providing no practical guidelines.

\begin{figure}[t]
	\begin{center}
		\scalebox{0.75}{
			\begin{tikzpicture}
				\node[punkt,left= of dummy, fill=lightgray, minimum width=80pt,draw=black] at (3,-3.0) (a) {\textsf{Mathematical model}};

				\node[punkt,fill=lightgray,minimum width=80pt,draw=black] at (-3.75,0) (c) {\textsf{Architectural choices}};
				\node[punkt, fill=lightgray,minimum width=80pt,draw=black]at (3.75,0) (d) {\textsf{Datasets}};

				\def\myshift#1{\raisebox{0.6ex}}
				\draw [>={Stealth[length=8pt,round]},<->, thick, bend right, postaction={decorate,decoration={text along path, raise=-0.50cm, text align=center,text={|\sffamily\myshift|Theoretical analysis}}}] (c.south) to   (a.west);

				\draw [>={Stealth[length=8pt,round]},<->, thick, bend right, postaction={decorate,decoration={raise=0.05cm, text along path,reverse path,text align=center,text={|\sffamily\myshift|Domain knowledge}}}] (a.east) to  (d.south);

				\draw [>={Stealth[length=8pt,round]},<->, thick, postaction={decorate,decoration={raise=0.05cm, text along path,text align=center,text={|\sffamily\myshift|Experiments}}}]   (c) to  (d);

				\node[punkt,left = of c, fill=ForestGreen!50,minimum width=85pt,draw=black] (i) {\textsf{Implementations}};
				\node[punkt, fill=ForestGreen!50,minimum width=85pt,draw=black] at (-8.1,-1.5) (iiiii) {\textsf{Baselines}};

				\node[punkt,right = of d, fill=ForestGreen!50,minimum width=85pt,draw=black] (ii) {\textsf{Benchmarks}};
				\node[punkt, fill=ForestGreen!50,minimum width=85pt,draw=black]  at (7.6,-1.5) (iii) {\textsf{Evaluation}};
				\node[punkt, fill=ForestGreen!50,minimum width=85pt,draw=black] at (7.6,-3.0)  (iiii) {\textsf{Exploration}};

				\node[lightgray] at (0.25,-1.35) (dojo) {\huge \textit{The Dojo} \begin{CJK}{UTF8}{min}道場\end{CJK}   };

			\end{tikzpicture}}
		\vspace{-5pt}
	\end{center}
	\caption{Proposal for a better alignment of theoretical and practical research within the graph machine learning community. We propose the tight interaction and iterative refinement of mathematical models and architectural choices via rigorous experimental evaluations supported by state-of-the-art baseline implementations, benchmarks, evaluation pipelines, and visual exploration tools. \label{fig:thecycle}}
\end{figure}
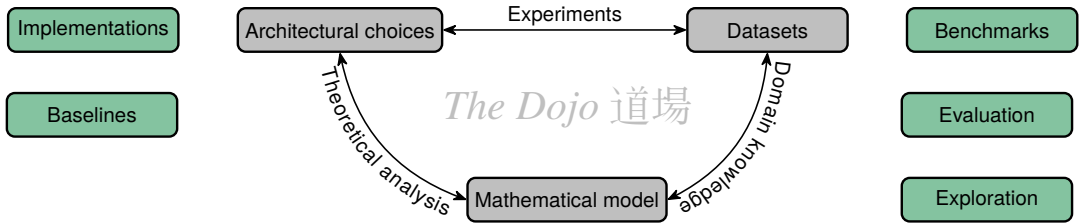
Hence, it is essential to adapt new theoretical results to domains where GNNs are frequently used, e.g., the molecular domain~\citep{Duv+2023} or combinatorial optimization~\citep{Cap+2021}. Moreover, in practice, using certain engineering tricks or folklore knowledge is common, e.g., using a specific normalization layer in an application domain. Therefore, it is essential to incorporate these choices into theory, understanding why they work in practice and how to improve them potentially.

Moreover, to quickly disseminate state-of-the-art, theoretically principled GNN architectures to real-world applications, providing efficient, easy-to-use implementations of such architectures is essential. \citet{Wan+2023} took a first step in this direction by providing open-source implementations of recently proposed expressive GNN architectures. Hence, it is crucial to push such initiative further by establishing a library used by the community, possibly extending \textsc{PyTorch Geometric}~\citep{Fey+2019} or \textsc{DGL}~\citep{Wan+2019}, making such architectures readily available and easily benchmarked for practitioners.

Similarly, theoretical papers often need more thorough experimental evaluations. They are mostly evaluated on small, out-of-date benchmark datasets, and often, their hyperparameters are not tuned sufficiently. Therefore, it is often unclear if they perform better than tuned state-of-the-art GNNs. Hence, it is essential to establish proper experimental pipelines and evaluation protocols for newly proposed theoretically-principled architectures. Alongside establishing solid evaluation protocols, it is important to establish (synthetic) benchmark datasets to investigate the effect of expressivity and its connection to generalization and optimization in detail. A first step in this direction was recently made by introducing the \textsc{Brec} dataset~\citep{Wan+2023b}, which still lacks diversity in graph structure.

In addition, with the emergence of large-language models (LLMs), several recent works tried using them for tasks such as node/graph classification or graph generation~\citep{Che+2024,Fat+2023}. However, their application remains mostly ad-hoc, and it is unclear when they outperform GNNs or help GNNs make better predictions. 

\subsection{Challenges}

We derive the following challenges from the above to better align theory with practical needs.

\paragraph{Challenge V.1: Unifying practical studies of theoretically principled GNN architectures.} We have argued that expressiveness, generalization, and optimization are interrelated aspects necessitating a more holistic treatment. We believe this should also be the case when studying these aspects from a practical, experimental perspective and suggest establishing a ``Theo-practical Dojo'' inspired by the work of~\cite {joshi2023ontheexpressive} on geometric graphs. The primary objective is to guarantee a unified and standardized experimental comparison of GNN architectures designed to guide theoretically grounded considerations and foster controlled, comparative studies. Accordingly, the dojo would consist of pivotal benchmarks extending beyond graph discrimination and encompassing tasks, including predicting relevant graph properties. Other than exposing standard protocols for training and evaluation in the spirit of~\citep{hu2020ogb}, to jointly elicit aspects related to expressiveness, generalization, and optimization, proposed tasks would cover families of graphs with diverse structural characteristics, various types of training-test distribution shifts and training datasets of different scales. The dojo should allow controlling for and contrasting the complexity of methods in comparison, regarding the number of learnable parameters and their empirical running time.

Finally, the dojo could be extended with a tool to visually and interactively explore the hidden representation space of models in comparison. We envision the tool would
enhance comprehension of the relationship between separation power and generalization. Overall, from the analyses supported by the dojo, researchers would obtain insights on their approaches, such as architectural ``blind spots,'' performance gaps, and scaling difficulties, in a way to informatively guide follow-up research refining their mathematical models, see~\Cref{fig:thecycle}.

\paragraph{Challenge V.2: A library of state-of-the-art, theoretically-guided GNN implementations.} To support the development of the above-outlined dojo, it is crucial to have an ideally large set of well-maintained and documented implementations of state-of-the-art theoretically-guided GNN architectures available, possibly building on existing GNN implementation libraries such as \textsc{PyTorch Geometric}~\citep{Fey+2019} or \textsc{DGL}~\citep{Wan+2019}. Moreover, the dojo's evaluation pipeline and available datasets should be integrated into such a library. In addition, we propose to provide one-click reproducible baseline results to compare newly proposed architectures to existing ones easily and ensure fair comparisons.

\paragraph{Challenge V.3: Adapting theoretically-guided GNN architectures to domain knowledge.} The challenges outlined in~\cref{sec:expressivity,sec:generalization,sec:optimization} mainly deal with the derivation of a general theory of graph machine learning. However, to quickly disseminate the theory to practice and leverage it by domain experts, it needs to be adapted to the specific needs of specific domains. Hence, we propose to identify key application domains for graph machine learning, e.g., molecular property prediction or combinatorial optimization, and, together with domain experts, work out their specific requirements. For example, it is essential to understand which kind of graph structures arise in particular domains or what engineering tricks, such as normalization layers, are currently leveraged. Building on such a requirements list, we propose to adapt and refine the mathematical theory and theoretical results. In addition, we also suggest investigating the potential of Neural Architecture Search~\citep{Col+2023,Els+2019} for GNNs to automatically adapt existing architectures to meet specific requirements. It would also be interesting to comparatively study architectural solutions found by NAS and those emerging by the theoretical studies proposed in this paper.

\paragraph{Challenge V.4: Improving graph machine learning in different learning paradigms.} Most of the graph machine learning works we have cited here study supervised graph machine learning, theoretically and empirically. Nonetheless, many other learning paradigms impact graph machine learning and other areas, e.g., self-supervised learning, generative modeling, transfer learning, foundation models, few-shot learning, meta-learning, reinforcement learning, and planning. Advances in graph machine learning within these different learning paradigms have found substantial empirical impact in several cases. Graph learning research has, for instance, fostered interesting advances in the area of AI agents and planning~\citep{sthalberg2021learning,deac2020graph} or has, otherwise, achieved noteworthy results when in combination with the aforementioned learning approaches, see, e.g., chip design using GNNs and reinforcement learning~\citep{mirhoseini2021graph}, self-supervised learning for 3D molecular property prediction~\citep{godwin2022simple}, and transfer learning from supervised pre-trained molecular models~\citep{ying2021transformers}. However, the success of some of these different learning paradigms in graph machine learning has lagged significantly behind their success in other areas, such as natural language processing, computer vision, and audio processing. For instance, self-supervised learning on pure graph data (without, e.g., the task-specific molecular 3D information as in~\citet{godwin2022simple}) has had much less empirical impact than in NLP and CV. Also, generative models for graphs are much weaker than those of other domains; sequence models that generate molecules via string representations are easier to use and often outperform graph generative models~\citep{flam2022language}, so sequence models are frequently used in practice for drug-design applications. Outside of graph machine learning, each learning paradigm has rich theory and specific methods that enable their success. Further theoretical and empirical study into connecting graph machine learning with these different learning paradigms could be very impactful.

\paragraph{Challenge V.5: The principled application of LLMs for GNNs.}

Nowadays, there is a large set of work on mixing text and molecule/protein/crystals modalities, e.g., performing protein retrieval based on their text descriptions~\citep{Xu+2023}, editing molecular structures with text instructions~\citep{Liu+2022}, or generating geometric properties of crystal structures solely from fine-tuned LLMs~\citep{Gru+2024}. Pioneering works also propose using LLMs for general interactive reasoning and mining on graphs~\citep{Fat+2023,zhao2023graphtext}. Hence, a future challenge regarding the foundations of graph learning would be to understand these model capabilities precisely, with a specific focus on their limitations and failure cases. Another pressing challenge is applying LLM/GPT-style training on graphs in a principled manner, e.g., for autoregressive graph generation. Here, we need to traverse the graph canonically to represent the graph as a sequence of tokens. However, theoretically, this is challenging as it entails determining the orbit of each node (the structural role of each node in the graph). Hence, a challenge here is to devise ``approximate'' node traversal strategies~\citep{zhao2023graphgpt} that work well with LLMs-like training objectives or devise exact ones for practical, relevant graph classes, e.g., molecular graphs.

\section{Conclusion}
Here, we stressed the importance of a broader theory of graph machine learning. Concretely, we highlighted the importance of developing a more fine-grained theory of expressivity, relying less on the simple perspective of graph isomorphism testing and considering more practically relevant architectural parameters such as normalization layers and skip connections. In addition, we underlined the need for more work investigating the generalization and optimization aspects of graph machine learning, focusing on more realistic assumptions. Of course, the aspects of expressivity, optimization and generalization are closely linked, and this \emph{interplay} warrants further understanding. Finally, we stressed the importance of aligning this more balanced theory of graph machine learning with practical needs, e.g., by considering expert knowledge. By investigating the challenges we have introduced, we believe that our research community will be able to rethink the pillars of expressiveness, generalization, and optimization in a more holistic and cohesive manner that is more directly informed by practical and domain considerations. We hope that this paper presents a valuable handbook of directions for developing a more realistic and balanced theory of graph machine learning and that its insights will help spur novel research results and avenues in the future.

\section*{Acknowledgements}
Christopher Morris is partially funded by a DFG Emmy Noether grant (468502433) and RWTH Junior Principal Investigator Fellowship under Germany’s Excellence Strategy. Nadav Dym is supported by Israeli Science
Foundation grant No.~272/23. Haggai Maron is the Robert J. Shillman Fellow and is supported by the Israel Science Foundation through a personal grant (ISF 264/23) and an equipment grant (ISF 532/23). Fabrizio Frasca is funded by the Andrew and Erna Finci Viterbi Post-Doctoral Fellowship. Ron Levie is supported by the Israel Science Foundation (grant No.~1937/23). Derek Lim is funded by an NSF Graduate Fellowship. Michael Bronstein is supported by the EPSRC Turing AI World-Leading Research Fellowship
No.~EP/X040062/1 and the EPSRC AI Hub for Mathematical Foundations of Intelligence No. EP/Y028872/1. Martin Grohe acknowledges funding from the European Union (ERC AdG SymSim, 101054974). Stefanie Jegelka acknowledges funding from NSF award CCF-2112665 (TILOS AI Institute) and the Alexander von Humboldt Foundation.

\bibliography{references}
\end{document}